\documentclass{llncs}
\usepackage{amssymb}
\usepackage{amsmath}
\usepackage{amsfonts}
\usepackage{algorithm}
\usepackage{makeidx}
\usepackage{graphicx}
\usepackage{epstopdf}
\usepackage[footnotesize]{subfigure}
\usepackage[usenames,dvipsnames]{color}
\usepackage{multirow}
\usepackage{hyperref}
\usepackage{url}
\usepackage{tikz}

\newcommand{\RR}{{\mathbb{R}}}
\newcommand{\ZZ}{{\mathbb{Z}}}

\newcommand{\BO}{{\mathbf{\Omega}}}

\newcommand{\BP}{{\mathbf{P}}}

\begin{document}
\title{Convolutional Neural Network on Semi-Regular Triangulated Meshes and its Application to Brain Image Data }
\author{Chaoqiang Liu$^1$,  Hui Ji$^1$, Anqi Qiu$^1$}

\institute{
$^1$ National University of Singapore, Singapore\\
\vspace{0.3cm}
%\red{\tt bieqa@nus.edu.sg}
}

\maketitle

\begin{abstract}
We developed a convolution neural network (CNN) on semi-regular  triangulated meshes whose vertices have 6 neighbours. The key blocks of the proposed CNN, including convolution and down-sampling, are directly defined in a vertex domain. By exploiting the ordering property of semi-regular meshes, the convolution is defined on a vertex domain with strong motivation from the spatial definition of classic convolution.
Moreover, the down-sampling of a semi-regular mesh embedded in a 3D Euclidean space can achieve a down-sampling rate of 4, 16, 64, etc. We demonstrated the use of this vertex-based graph CNN for the classification of mild cognitive impairment (MCI) and Alzheimer's disease (AD) based on 6767 MRI scans of the Alzheimer's Disease Neuroimaging Initiative (ADNI). We compared the performance of the vertex-based graph CNN with that of the spectral graph CNN.
\end{abstract}

%\vspace{-0.8cm}
\section{Introduction}
Machine learning has been widely used as one crucial technique for medical image segmentation, registration, disease prediction and classification, in which image data are sampled on a Euclidean equi-spaced grid. However, the geometry of human organs is in general very complex, which  characterizes the intrinsic properties of anatomy and physiological functions of the organs. For instance, myocardial  contraction flows along the wall of the heart. Cortical thickness, representing the depth of the cortical ribbon, is related to the nature of the convoluted gyri and sulci of the cortex. Cortical thickness is thicker on cortical gyri but thinner on cortical sulci. Hence, it is preferred to represent brain image data in terms of its geometry that can be expressed as meshes embedded in the Euclidean space of the brain image. It has been demonstrated that geometric structure relating image data did introduce useful information in machine learning based methods for
 disease diagnosis (e.g., \cite{YangQiu2013,davatzikos_biopsy_2008,apostolova_brain_2006,anqi_adni_2009}).
 
 In recent years, deep learning has been one of prevalent machine learning techniques to tackle a wide range of image-related applications. Among many architectures of deep neural networks,  
\emph{convolutional neural network} (CNN) received a great attention for its successes in computer vision (e.g. \cite{simonyan2014very,redmon2016you,xie2012image,Szegedy2015,Krizhevsky2012,he2016deep}), medical imaging and diagonosis (e.g.,  \cite{shin2016deep,milletari2016v,esteva2017dermatologist,litjens2017survey}).
 The main blocks to build a CNN include  convolution with localized filters, non-linear 
activation function, and pooling. In CNN, these three blocks are sequentially concatenated to model highly non-linear intrinsic patterns of training data and output the features for targeting applications.   Most of these CNNs are developed for modeling image data defined on equi-spaced regular grids. The generalization of such CNNs to image data defined on the meshes embedded in a higher dimensional Euclidean space is non-trivial, especially for the localized convolution and pooling operations.

\subsection{CNN on general graphs}
One might view meshes as a special class of graphs. There have been several works on generalizing the CNN for modeling data on general graphs; (e.g.,  \cite{bruna2013spectral,Defferrard2016,henaff2015deep}). Based on spectral graph theory, Henaff et~al. \cite{henaff2015deep} proposed a CNN for graph-structured data, in which  convolution is defined as a diagonal multiplicative operation in graph Fourier transform derived from a normalized graph Laplacian. The localization of the convolution is imposed by regularizing those diagonal entries with a smoothness prior. To avoid the computation of a graph Laplacian and
have a convolution with better localization,  Defferrard et~al. \cite{Defferrard2016} introduced Chebyshev polynomial approximation such that the resulting convolution operator is a polynomial of the adjacency matrix of a graph. Kipf and Welling \cite{kipf2016semi} further simplified the approximation using the linear polynomial of the adjacency matrix of a graph and applied the CNN for semi-supervised learning.

Nevertheless, the convolution built on the polynomials of the adjacency matrix is very  different from classic convolution on equi-spaced grids. Consider such a generalized convolution derived from the polynomial degree $k$. Then, it is parameterized by totally $(k+1)$ parameters. The support of such a convolution, i.e., the number of the vertices it covers for each shift, is  $3k^2+3k+1$ vertices for a regular triangular mesh. The idea of classic CNN for modeling images on equi-spaced grids is that it uses small filters to collect as much local information as possible, and then gradually increase the filter width and down-size the features to represent more global and high-level information.  A convolution that covers a large number of vertices might lose important local features which are helpful for modeling.

As mentioned above, how to down-size the feature is also an essential operation for CNN to abstract more high-level information. Such a down-sizing operation happens in both pooling and convolution with stride $>1$. The  graph coarsening procedure used for the pooling in \cite{Defferrard2016} is implemented by calling a weighted graph cut method  \cite{dhillon2007weighted}. From the coarsest to the finest level, fake vertices, i.e. disconnected vertices, are added to pair with the singletons such that each vertex has two children. The fake vertices artificially increase the dimensionality and thus the computation cost even though the number of singletons from multilevel clustering algorithms may not be large.

\subsection{CNN on semi-regular triangular meshes}
In this study, we proposed a vertex-based CNN approach for modeling image data defined on semi-regular triangular meshes that are well structured in terms of connectivities, e.g., the connectivity of most vertices  is $6$. A semi-regular triangular mesh 
has certain similarities to equi-spaced grids in the Euclidean space. When image data are defined on a semi-regular triangular mesh, a direct call of a generic CNN for a graph certainly  is sub-optimal, as it discards specific connectivity properties of the mesh. Indeed, the  connectivity property of a triangular mesh enables us to better mimic convolution and down-sizing operation such as to avoid the issues encountered in the CNN defined on a graph discussed in the previous section.

The key blocks of the CNN proposed in this paper, especially convolution and down-sizing, are directly defined in a vertex domain. By exploiting the ordering property of semi-regular meshes, the convolution is defined on a vertex domain with strong motivation from the spatial definition of classic convolution.
 Moreover, the down-sampling of a semi-regular mesh was efficient. The down-sampling of a semi-regular mesh embedded in a 3D Euclidean space can achieve a down-sampling rate of 4, 16, 64, etc. We demonstrated the use of this vertex-based graph CNN for the classification of mild cognitive impairment (MCI) and Alzheimer's disease (AD) based on 3169 image datasets of the Alzheimer's Disease Neuroimaging Initiative (ADNI). We compared the performance of the vertex-based graph CNN with that of the spectral graph CNN \cite{Defferrard2016}.

%which exploits the geometrical regularity of   such meshes, with a particular focus on compact localization of convolution and efficient down-sampling.  
 
 %whose vertex has .  Meshes are well structured in terms of connectivities, eg., the connectivity of most vertices  is $6$ for a semi-regular mesh. It geometrical and topological properties is quite similar  to equi-spaced grids in many aspects. In this paper,  we introduced a vertex based CNN on semi-regular triangulated meshes 

% Based on spectral graph theory, there have been works on generalizing CNN for signal on  arbitrarily structured graphs \cite{bruna2013spectral,henaff2015deep}, which treats convolution as diagonal operations defined on graph laplacian. To avoid computation of Graph Laplacian, the adjacency matrix based Chebyshev polynomial approximation is employed \cite{defferrard2016,Kipf2017}, in which the induced convolution is also localized when the connectivity of graph vertices is local. 

%A mesh can be viewed as the discretization of a manifold. There are several works on generalizing neural network to discrete a manifold \cite{masci2015geodesic,boscaini2016learning,monti2017geometric}. 
%These methods are also based on spectral graph theory, and their focus is on the construction of  appropriate local metric for shape analysis, not for signal analysis on manifold. In addition, 

\section{Methods}

\subsection{Convolution in the vertex domain}
Consider a signal $f$ defined on an equi-spaced grid $\{k\}_{k\in \ZZ^2}$, and a finite filter $h$ supported on a finite set $\BO\subset \ZZ^2$. The convolution is then defined by
\begin{equation}\label{eqn:conv}
(f\otimes h)[m]=\sum_{n\in \ZZ^2}f[n]h[m-n]=\sum_{n\in m-\BO}f[n]h[m-n]
\end{equation}
It can be seen that at the vertex $m$, the value of $f\otimes h$ is indeed  weighted average of $f$ over the neighbors of the vertex $m_0$, whose weights are given by $h$ and neighbors are determined by $\BO$. In the following, we generalize such a concept to a semi-regular triangular mesh whose vertex in general has $6$ neighbors.

\begin{figure}
	\centering
	\includegraphics[width=0.4\linewidth]{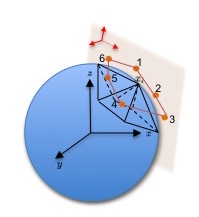}
	\caption{The ordering of the 1-ring neighbors. The mesh, $T$, is represented in black. The sphere (blue) approximates the neighbors of vertex $x_i$ and the tangent plane at $x_i$ is in orange. The points colored in orange are the projected points of the neighbors of vertex $x_i$ on its tangent plane. }
	\label{fig:projection}
\end{figure}

We begin with a semi-regular triangular mesh 
$$T=(\{x_i\},\{\Sigma_{ijk}\}),i,j,k\in \{1,\dots,N\},$$
where $\{x_i\}$ denotes the set of vertex coordinates. $\{\Sigma_{ijk}\}$ is the set of simplices and $N$ represents the total number of vertices on the mesh. Each simplex $\Sigma_{ijk}$ is a three tuple of points $(i,j,k), i,j,k \in \{1,\dots,N\}$ that specifies the vertices forming a triangular face, i.e., all three vertices $x_i, x_j$ and $x_k$ are one-ring neighbours of each other. 
If  $T$ is a semi-regular mesh, then for each vertex, $x_i$, the set of its neighboring vertices can be denoted as $\BP_i\subset \{x_i\}$ :
$$
\BP_i=\{P[i,1],P[i,2],\ldots, P[i,6]\}.
$$
The ordering of these vertices for this convolution is not straightforward. Fig. \ref{fig:projection} illustrates how these vertices are ordered in this study. We first define a sphere, $\mathcal{S}_i$ (blue in Fig. \ref{fig:projection}), that passes $x_i$ and approximates the mesh formed by $\BP_i$. The tagent plane (orange in Fig. \ref{fig:projection}) of $x_i$ on the mesh, $T$, is defined as the tagent plane of $x_i$ on the sphere, $\mathcal{S}_i$. The $xyz-$coordinate of the tagent plane of $x_i$ (red in Fig. \ref{fig:projection}) is the translation and rotation of the coordinate of $\mathcal{S}_i$ (black in Fig. \ref{fig:projection}). We then order these six vertices in a closewise sequence,  where $P[i,1]$ is defined as the vertex whose projection is the closest to the x-axis of the tagent plane of the vertex, $x_i$.

Consider a $1$-ring filter $h\in \RR^7$:
$$
h=[h[0],h[1],h[2],\ldots h[6]]^\top.
$$
Then, at the vertex $x_i$, the value of a signal $f$ defined on $T$  convolved by $h$ is  defined by
\begin{equation}\label{eqn:mconv}
(f\otimes h)[i]=\sum_{j=0}^6 \widetilde{f}[i,j]h[j],
\end{equation}
where $\widetilde{f}[i,j]$ denote the value of $f$ at the vertex $P[i,j]$ and $P[i,0]=x_i$. 

%\begin{figure}
%	\centering
%	\includegraphics[width=1\linewidth]{../figure/CNN.jpg}
%	\caption{One layer of the geometric neural network. Panel A illustrates one layer of the neural network with the convolution with a filter shifting over th mesh (indicated in red), ReLU, and pooling. Panel B illustrates the pooling operation as all vertices and edges in red are removed for the downsampling.}
%	\label{fig:layer}
%\end{figure}

In a matrix form, we define a matrix $D\in \RR^{7,N}$ as 
$$
D=\left[
\begin{array}{ccccc}
\widetilde f[1,0] & \cdots & \widetilde f[i, 0] & \cdots &  \widetilde f[N,0] \\
\widetilde f[1,1] & \cdots & \widetilde f[i,1] & \cdots
& \widetilde f[N,1]  \\
\vdots & \vdots & \vdots & \vdots & \vdots \\
\widetilde f[1,6] & \cdots & \widetilde f[i,1] & \cdots
& \widetilde f[N,6] 
\end{array}
\right]
$$
Then, the convolution defined in Eq. \eqref{eqn:mconv} can be expressed in the form of a matrix multiplication:
\begin{equation}
\label{eqn:matrixconv}
f\otimes h: f\in \RR^N \rightarrow h^\top D \in \RR^N.
\end{equation}
For the vertex in $T$ with valence $<6$, whose corresponding column has non-defined entries. Analogous to classic convolution for finite signals, we can define the values of these entries using boundary extension. For example, assigning 0 to these entries which is the same as zero padding boundary extension in classic convolution.

By the same procedure, we can define $2$-ring convolution and more.
Consider a $k$-ring convolution, it is parameterized by totally $3k(k+1)+1$ parameters, and its support also covers the same number of vertices. This is consistent with the behavior of classic convolution on equi-space grids. Such localization property enables CNN to extract very local features of the data on semi-regular triangular meshes, the same as what CNN is doing on equip-spaced grids. Such  ring-type convolution also has been exploited in wavelet transform for surface processing \cite{dong2016multiscale}.

% In contrast, Chebychev polynomial based convolution with a $k$-tap filter will cover such a number of vertices. See Fig.~\ref{fig:conv} for an illustration.
%\begin{figure}
%	\centering
%	\includegraphics[width=1\linewidth]{./fig/work_flow.jpg}
%	\caption{Architecture of CNN on mesh.}
%	\label{fig:conv}
%\end{figure}

\subsection{Rectified Linear Unit and Pooling}
For CNN, there are many types of non-linear activation function. The activation function is a map from $\RR$ to $\RR$, which does not involve any geometrical property of the underlying structure. In our proposed CNN for image data on a semi-regular mesh, we adopt the well-known \emph{rectified linear unit} (ReLU):
$$
f(x)=\max\{0, x\},\quad x\in \RR.
$$ 

In addition to convolution and ReLU, another important block is \emph{pooling}, which can be viewed as a non-linear or linear down-sampling operation. The pooling enables us to reduce the size of representation and thus to reduce the number of parameters, which helps memory usage, computational efficiency and  over-fitting controlling. The pooling is done by either taking the maximum or taking the average of the  neighbors of those vertices lying on the coarser grid/mesh. The key for defining a pooling operation on a mesh is about how to define a hierarchical triangular mesh:
$$
\{T^{(0)},T^{(1)},T^{(2)},\cdots,T^{(L)}\}
$$
such that the vertices of $T^{(j+1)}$ contain all vertices of $T^{(j)}$ and new vertices, and $T^{(L)}$ is the original mesh $T$ on which  image data is defined.

\begin{figure}
	\centering
	\includegraphics[width=0.9\linewidth]{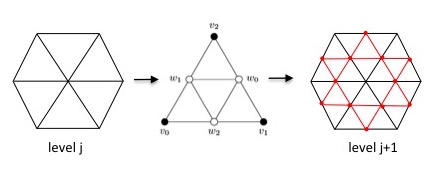}
	\caption{The subdivision scheme of a triangulated mesh.}
	\label{fig:sub}
\end{figure}

As the design of the proposed CNN mainly aims at modeling brain image data, we first generate a hierarchical semi-regular triangular mesh such that the mesh in the finest scale is the mesh extracted from image data. There are many approaches for generating a hierarchical triangular mesh and we adopt the one used \cite{lounsbery1997multiresolution} which recursively uses subdivision scheme to generate new vertices. Consider a triangle in the mesh $T^{(j)}$ with 3 vertices $(v_0,v_1,v_2)$. Then, the triangle is subdivided into $4$ smaller ones by $(w_0,w_1,w_1)$, which are the midpoints of three edges of this triangle. The four new triangles are given by
$$
(v_0,w_2,w_1),\quad (v_1,w_0,w_2),\quad (v_2,w_1,w_0),\quad
(w_0,w_1,w_2).
$$
See Fig.~\ref{fig:sub} for an illustration.

%
%\begin{figure}
%	\centering
%	\begin{tabular}{c@{\hspace{5pt}}c}
%		\begin{tikzpicture}
%		\draw (1.5,0) -- (2.25,1.3);
%		\draw (1.5,0) -- (0.75,1.3);
%		\draw(2.25,1.3) -- (0.75,1.3);
%		\draw (0,0) node[circle,fill=black,inner sep=0pt,minimum size=6pt,label=below:{$v_0$}]{$$}
%		-- (1.5,0) node[circle,draw=black, fill=white, inner sep=0pt,minimum size=6pt,label=below:{$w_2$}]{$$}
%		-- (3,0) node[circle,fill=black,inner sep=0pt,minimum size=6pt,label=below:{$v_1$}]{$$}
%		-- (2.25,1.3) node[circle,draw=black, fill=white, inner sep=0pt,minimum size=6pt,label=right:{$w_0$}]{$$}
%		-- (1.5,2.6) node[circle,fill=black,inner sep=0pt,minimum size=6pt,label=above:{$v_2$}]{$$}
%		-- (0.75,1.3) node[circle,draw=black, fill=white, inner sep=0pt,minimum size=6pt,label=left:{$w_1$}]{$$}
%		-- (0,0); 
%		\end{tikzpicture} &
%			\includegraphics[width=0.6\linewidth]{../figure/mra_mesh.pdf}
%			 \\
%		(a) & (b)
%	\end{tabular}
%\label{fig:sub}
%\caption{(a) The subdivision scheme of an triangle; (b) hierarchy semi-regular triangular mesh for sphere}
%\end{figure}
%\begin{remark}
%	In the case that image data is defined on semi-regular meshes for sphere. the midpoint of any pair of vertices
%	is defined as the midpoint of spherical geodesic arc connecting these two vertices.
%\end{remark}

Starting with an initial mesh at the coarsest level, recursively applying the subdivision scheme above leads to a hierarchical semi-regular triangular mesh.
$$
T^{(0)}\rightarrow T^{(1)}\rightarrow T^{(2)}\rightarrow \cdots.
$$
The vertex number of the mesh at each level is $4$ times that of the mesh at the next coarse level.
It can be seen that for any vertex $x_i^{(j)}$ at the $j$-th level mesh $T^{(j)}$, it remains in the $(j+1)$-th level mesh
$T^{(j+1)}$, and all its $1$-ring neighbors are $6$ new vertices not in $T^{(j)}$. Then, for vertex $x_i^{(j)}$ at the $j$-th level mesh $T^{(j)}$, let $\Omega_{i}^{(j+1)}$ denote the set of this vertex and all its $1$-ring neighbors.
 Then,
the pooling operator with stride $2$ is defined as
$$
\begin{array}{ll}
\text{Mean pooling}: & f[i]^{(j)} \longrightarrow \frac{1}{7}\sum_{r\in \Omega_{i}^{(j+1)}}
f[r]^{(j+1)};\\
\text{Max pooling}: & f[i]^{(j)} \longrightarrow \max_{r\in \Omega_{i}^{(j+1)}}
f[r]^{(j+1)},
\end{array}
$$
where $f[i]^{(j)}$ denotes the value at the vertex $x_i$ in the $j$-th level mesh $T^{(j)}$.
Similarly, we define the pooling operator with stride $2$ (and more) by running the same procedure on the vertices and all its $1$-ring and $2$-ring neighbors (and more) in the next finer level mesh.

\subsection{CNN on a semi-regular mesh}

\begin{figure}
	\centering
	\includegraphics[width=0.9\linewidth]{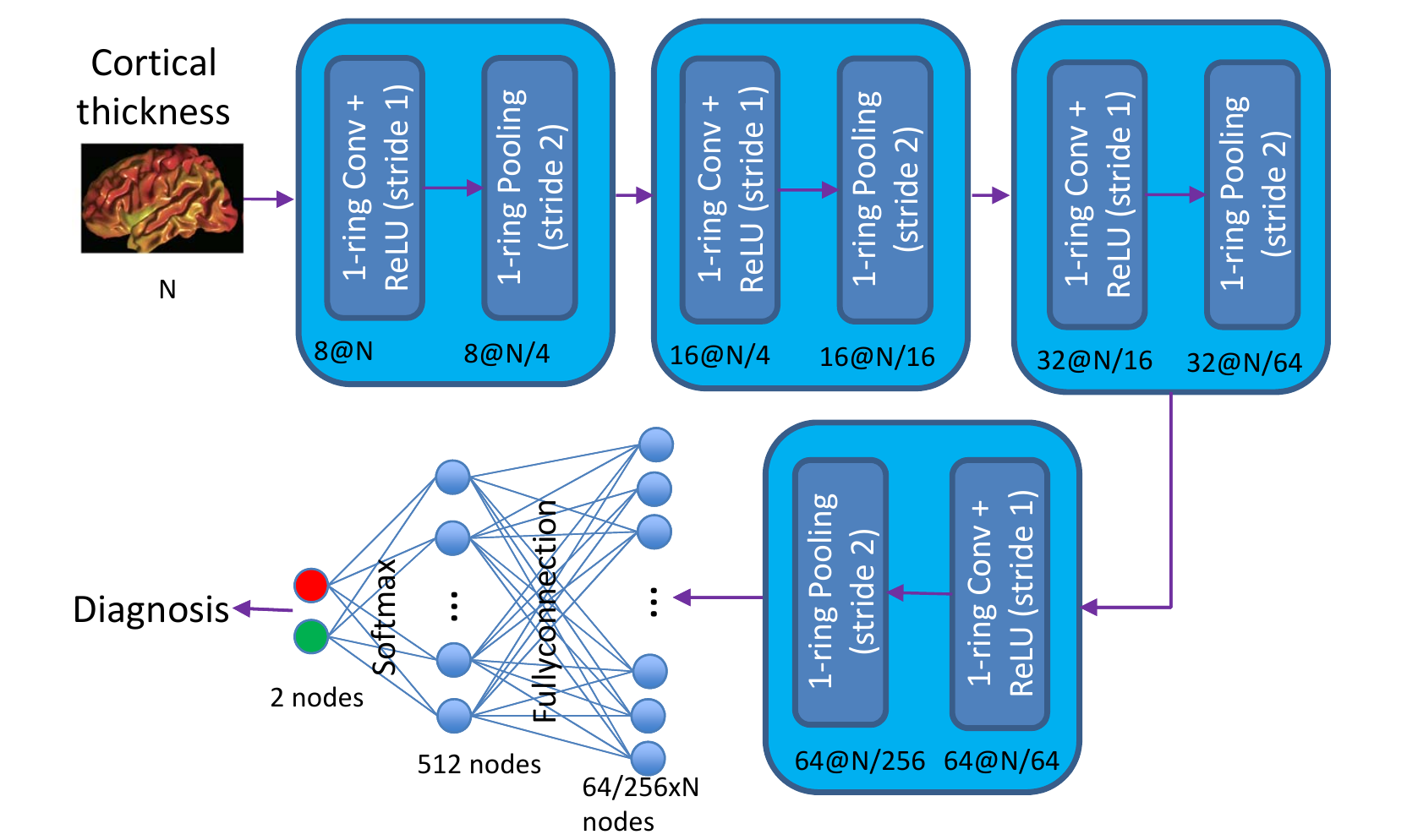}
	\caption{The architecture of the proposed CNN on  a semi-regular triangulated mesh. The four layers respectively include 8, 16, 32, and 64 filters. $N$ represents the number of vertices on the input mesh.}
	\label{fig:network}
\end{figure}

Based on main ingredients presented in the previous section, 
we propose a vertex-based CNN for analyzing image data defined on a semi-regular mesh,  which is analogous to classic CNN for image data defined on equi-spaced grids.
The CNN is composed of totally $L+1$ connected blocks $M^{(1)},\ldots,M^{(L)}, M^{(L+1)}$. The first $L$ blocks are the blocks  for feature extraction. Each block  contains three sequentially concatenated layers: (1) a convolution layer with multiple $1$-ring convolutions; (2) a ReLU layer;  (3)  a pooling layer with stride 2 that uses mean pooling:
$$
M^{(\ell)}: \text{input}\rightarrow \text{ Convolutions}\rightarrow \text{ReLU}\rightarrow
\text{Mean Pooling}\rightarrow \text{Output}
$$
The last block $M^{(L+1)}$ is the classification layer using the features extracted from the previous blocks,

In our implementation of the proposed CNN for classifying brain image data. It has  4 feature extraction blocks and 1 classification block.
The numbers of the convolution filters in these  blocks are [8,16,32,64] respectively.
The  classification layer is implemented using a fully connected layer with $512$ nodes and  with a softmax output.
Fig. \ref{fig:network} shows the architecture of the proposed CNN.
The $1$-ring convolution operation can be implemented in a matrix multiplication. The other layers can be implemented using the standard procedure. In our implementation, the CNN is trained using the Adam optimization algorithm. We implemented this CNN on a semi-regular triangulated mesh in Tensorflow. The code is available at the website~\footnote{ \href{url}{http://bioeng.nus.edu.sg/cfa/download/GCNN.zip}}

\section{Results}%\vspace{-0.2cm}
\label{sec:expts}

\subsection{MRI Data and Analysis}
\noindent{\bf ADNI cohort.}   This study employed the data from the ADNI-1 and ADNI-2 cohort. The ADNI-2 cohort only included 1149 subjects (400 cognitive normal (CON), 301 early MCI (EMCI), 187 late MCI (LMCI) and 261 AD). The ADNI-1 cohort only included 1013/3598 subjects/scans (243/1067 CON, 415/1515 MCI, and 355/1016 AD). The number of visits of each subject varied from 1 to 7 (i.e., baseline, 3-, 6-, 12-, 24-, 36-, and 48-month). At each visit, subjects were diagnosed as one of the four clinical statuses based on the criteria described in the ADNI-2 protocol (\href{ulf}{http://adni.loni.usc.edu}). The general diagnostic criteria for early and late MCI were the same except LMCI subjects had a lower cut-off point for logical memory II subscale from Wechsler Memory Scale. The demographic and clinical information of subjects from ADNI-2 and ADNI-2 are provided in Table \ref{tab1} and Table \ref{tab1:ADNI1}.

\begin{table}[!htbp]
	\centering 
	\caption{Demographic and clinical information of the ADNI-2 cohort based on MRI scans. \label{tab1}}
	\begin{tabular}{|l|c|c|c|c|}
		\hline 
		\textbf{} & \textbf{   CON   } & \textbf{   EMCI   } & \textbf{   LMCI   } & \textbf{   AD   } \\
		\hline 
		\hline
		Number of subjects$^{*}$ & 400 & 301 & 187 & 261 \\
		Number of scans & 1122 & 865 & 595 & 587 \\ 
		Female/Male & 607/515 & 395/470 & 268/327 & 254/333 \\
		Age (Mean$\pm$SD) & 75.3$\pm$6.8 & 72.6$\pm$7.5 & 73.6$\pm$8.0 & 75.3$\pm$7.7 \\
		\hline 
	\end{tabular}
	\\ \begin{flushleft} \textbf{Abbreviations.} CON: Control normal; AD: Alzheimer's disease; MCI: Mild cognitive impairment; EMCI: Early MCI; LMCI: Late MCI. \\
	${*}$ The number of subjects for each group was based on the clinical status during the MRI acquisition visit. There were subjects who fall into 2 or more groups due to the conversion from one clinical status to another. \end{flushleft}
\end{table}

\begin{table}[!htbp]
	\centering 
	\caption{Demographic and clinical information of the ADNI-1 cohort based on MRI scans. \label{tab1:ADNI1}}
	\begin{tabular}{|l|c|c|c|}
		\hline 
		\textbf{} & \textbf{   CON   } & \textbf{   MCI   } & \textbf{   AD   } \\
		\hline 
		\hline
		Number of subjects$^{*}$ & 243 & 415 & 355 \\
		Number of scans & 1067 & 1515 & 1016 \\ 
		Female/Male & 493/574 & 525/990 & 443/573 \\
		Age (Mean$\pm$SD) & 76.8$\pm$5.3 & 75.9$\pm$7.3 & 76.3$\pm$7.2 \\
		\hline 
	\end{tabular}
%	\\ \begin{flushleft} \textbf{Abbreviations.} CON: Control normal; AD: Alzheimer's disease; MCI: Mild cognitive impairment; \\
%	${*}$ The number of subjects for each group was based on the clinical status during the MRI acquisition visit. There were subjects who fall into 2 or more groups due to the conversion from one clinical status to another. \end{flushleft}
\end{table}

%\noindent{\bf MRI Acquisition.} Structural T1-weighted MRI scans iwere acquired using either 1.5T or 3T scanners. For the 1.5T scanners, the imaging  protocol followed: repetition time (TR) = 2400 $ms$, minimum full echo time (TE), inversion time (TI) = 1000 $ms$, flip angle = 8$^\circ$, field-of-view (FOV) = 240 $\times$240 $mm^2$, acquisition matrix = 256$\times$256$\times$170 in the x-, y-, and z-dimensions, yielding a voxel size of 1.25$\times$1.25$\times$1.2 $mm^3$. For the 3T scanners, the imaging protocol was set to be: TR = 2300 $ms$, minimum full TE, TI = 900 $ms$, flip angle = 8$^\circ$, FOV = 260$\times$260 $mm^2$, acquisition matrix = 256$\times$256$\times$170, yielding a voxel size of 1.0 $\times$ 1.0 $\times$1.2 $mm^3$. 

\noindent{\bf MRI Data Analysis.} All T1-weighted images were segmented using FreeSurfer \cite{Fischl2002}. The processed images were quality checked based on the criteria listed on the website~\footnote{\href{url}{ https://surfer.nmr.mgh.harvard.edu/fswiki/FsTutorial/TroubleshootingData}}. We represented cortical thickness on the cortical surface generated by FreeSurfer. We employed a large deformation diffeomorphic metric mapping (LDDMM) algorithm \cite{Zhong2010,Du2011} to align individual cortical surfaces to the atlas and transferred the thickness of each individual subject to the atlas. 

As each subject may have multiple MRI scans, one at each visit, this study included all available T1-weighted images with good quality after processing. We used the clinical status at the MRI acquisition as the classification ground truth. For instance, a subject with multiple scans may have different clinical labels if he/she was from one clinical status to another over time. From 3365 scans from ADNI-2, we discarded 196 scans that missed demographic information or diagnosis labels of CON, EMCI, LMCI and AD, resulting in 3169 scans used in the following CNN analysis. From 3783 scans from ADNI-1, we discarded 185 scans that missed demographic information or diagnosis labels of CON, MCI, and AD, resulting in 3598 scans used below.

\subsection{Comparison with the Graph CNN}
In this experiment, we compared the computational cost and classification accuracy between the proposed vertex-based CNN and graph CNN \cite{Defferrard2016} based on the ADNI-2 data. The graph CNN \cite{Defferrard2016} incorporated 3 CNN layers with the number of filters of [8,16,32] respectively and a final fully connected layer with $128$ nodes. The convolution in the graph CNN was approximated using Chebyshev polynomial with the order of 3. The network parameters were trained with a mini-batch size of 64, an initial learning rate of $1e^{-3}$, a weight decay of 0.05, and a momentum of 0.9. During the training process, a $l_2-$norm regularization function of $5e^{-4}$ was applied on the weights of the final fully connected layer to prevent overfitting to the training data. This study employed the 10-fold cross-validation, where the scans from the same subject were assigned to the validation (or testing) to avoid the data leakage issue in the predictive model. We determined the parameters, such as the number of layers and the number of filters, and a learning rate, based on geometric mean (GMean=$\sqrt{SEN \times SPE}$, where  $SEN$ and $SPE$ respectively represent sensitivity (SEN) and specificity (SPE). We chose this measure because it not only maximized the accuracy on each of the two classes but also minimized the difference between the sensitivity and specificity, i.e., the balanced performance for both the positive and negative classes. 

We performed the same procedure as mentioned above for six classifiers, including CON vs. AD, CON vs. LMCI, CON vs. EMCI, EMCI vs. LMCI, EMCI vs. AD, and LMCI vs. AD. The experiments were run on Telsa M40 GPU (24GB memory). The computational time for each epoch of our vertex-based CNN and graph CNN \cite{Defferrard2016} was respectively 40 sec and 113 sec. Our proposed approach was $2.83$ times faster than the graph CNN. Table \ref{tab2} lists the classification accuracy, sensitivity, specificity, and GMean for each classifier. Our proposed vertex-based CNN was better performed than the graph CNN in most of the classifiers, including CON vs. AD, CON vs. LMCI, CON vs. EMCI, EMCI vs. AD, and LMCI vs. AD, except the EMCI vs. LMCI classifier. In addition, our vertex-based approach provided a relatively lower variability across all the four evaluation measures. These findings suggested that the proposed CNN is a fast computational model and has the potential to improve classification accuracy compared to the graph CNN \cite{Defferrard2016}.

\begin{table}[!htbp]
	\centering 
	\caption{Comparison between vertex-based and graph CNN in terms of accuracy (ACC), sensitivity (SEN), specificity (SPE), and geometric mean (GMean)  \label{tab2}}
\begin{tabular}{|c|c|c|c|c|c|c|}
	\hline 
	Model & Task & ACC(\%) & SEN (\%) & SPE (\%) & GMean(\%)\tabularnewline
	\hline 
	\hline 
	\multirow{6}{*}{Vertex-based CNN} & CON vs. AD & \bf{89.0 $\pm$ 0.6} & \bf{86.4 $\pm$ 1.1} & \bf{90.3 $\pm$ 1.1} & \bf{88.4$\pm$ 0.6}\tabularnewline
	\cline{2-7} \cline{3-7} \cline{4-7} \cline{5-7} \cline{6-7} \cline{7-7} 
	& CON vs. LMCI & \bf{73.3$\pm$ 1.1} & \bf{67.5$\pm$ 2.5} & \bf{76.4$\pm$ 1.7} & \bf{71.8$\pm$ 1.2}\tabularnewline
	\cline{2-7} \cline{3-7} \cline{4-7} \cline{5-7} \cline{6-7} \cline{7-7} 
	& CON vs. EMCI & \bf{67.9$\pm$ 2.6} & \bf{67.0$\pm$ 2.6} & \bf{68.7$\pm$ 2.5} & \bf{67.8$\pm$ 1.0}\tabularnewline
	\cline{2-7} \cline{3-7} \cline{4-7} \cline{5-7} \cline{6-7} \cline{7-7} 
	& EMCI vs. LMCI & 55.6$\pm$ 1.4 & 51.2$\pm$ 2.5 & 58.7$\pm$ 2.4  & 54.8$\pm$ 1.4\tabularnewline
	\cline{2-7} \cline{3-7} \cline{4-7} \cline{5-7} \cline{6-7} \cline{7-7} 
	& EMCI vs. AD & \bf{79.9$\pm$ 1.4} & \bf{75.2$\pm$ 1.7} & 83.1$\pm$ 1.7 & \bf{79.0$\pm$ 1.4}\tabularnewline
	\cline{2-7} \cline{3-7} \cline{4-7} \cline{5-7} \cline{6-7} \cline{7-7} 
	& LCMI vs. AD & \bf{65.4$\pm$ 1.3} & \bf{66.5$\pm$ 2.2} & 64.4$\pm$ 2.7 & \bf{65.4$\pm$ 1.4}\tabularnewline
	\hline 
	\multirow{6}{*}{Graph CNN} & CON vs. AD & 85.8$\pm$ 0.8 & 83.5$\pm$ 3.2 & 87.5$\pm$ 2.8  & 85.4$\pm$ 0.8\tabularnewline
	\cline{2-7} \cline{3-7} \cline{4-7} \cline{5-7} \cline{6-7} \cline{7-7} 
	& CON vs. LMCI & 69.3$\pm$ 2.2 & 65.6$\pm$ 7.6 & 72.0$\pm$ 5.4 & 68.5$\pm$ 3.0\tabularnewline
	\cline{2-7} \cline{3-7} \cline{4-7} \cline{5-7} \cline{6-7} \cline{7-7} 
	& CON vs. EMCI & 51.8$\pm$ 1.2 & 55.3$\pm$ 5.1 & 48.6$\pm$ 6.4  & 53.5$\pm$ 4.2\tabularnewline
	\cline{2-7} \cline{3-7} \cline{4-7} \cline{5-7} \cline{6-7} \cline{7-7} 
	& EMCI vs. LMCI & 60.9$\pm$ 2.2 & 52.5$\pm$ 8.8 & 67.8$\pm$ 9.8  & 59.1$\pm$ 1.4\tabularnewline
	\cline{2-7} \cline{3-7} \cline{4-7} \cline{5-7} \cline{6-7} \cline{7-7} 
	& EMCI vs. AD & 79.2$\pm$ 2.6 & 70.4$\pm$ 4.7 & 85.8$\pm$ 4.7  & 77.6$\pm$ 2.7\tabularnewline
	\cline{2-7} \cline{3-7} \cline{4-7} \cline{5-7} \cline{6-7} \cline{7-7} 
	& LCMI vs. AD & 65.2$\pm$ 1.6 & 62.6$\pm$ 5.2 & 68.0$\pm$ 6.6 & 65.3$\pm$ 1.4\tabularnewline
	\hline 
\end{tabular}
%\\ \begin{flushleft} \textbf{Abbreviations.} CON: Control normal; AD: Alzheimer's disease; MCI: Mild cognitive impairment; EMCI: Early MCI; LMCI: Late MCI; ACC: accuracy; SEN: sensitivity; SPE: specificity; GMean: adjusted geometric mean. \\
%	 \end{flushleft}
\end{table}

\subsection{Application to ADNI-1}
In this study, we applied the CON vs LMCI, CON vs AD, and LMCI vs AD classifiers obtained from the ADNI-2 cohort to the ADNI-1 cohort. Table \ref{tab:adni1} lists the classification accuracy for MCI and AD, which is comparable to those listed in Table \ref{tab2}. This suggests the robustness of the classifiers built based on the ADNI-2 cohort to the other dataset.

\begin{table}[!htbp]
	\centering 
	\caption{Classification to the ADNI-1 dataset in terms of accuracy (ACC), sensitivity (SEN), specificity (SPE), and geometric mean (GMean)  \label{tab:adni1}}
\begin{tabular}{|c|c|c|c|c|}
	\hline 
Task & ACC(\%) & SEN (\%) & SPE (\%) & GMean(\%)\tabularnewline
	\hline 
	\hline 
CON vs. AD & 88.9  & 82.3 & 95.2 & 88.5 \tabularnewline
CON vs. MCI & 67.7 & 55.8 & 84.6 & 68.7\tabularnewline
MCI vs. AD & 65.2 & 80.6 & 54.9 & 66.5\tabularnewline
	
		\hline 
	\end{tabular}
\end{table}

\section{Discussion}
This paper presented a vertex-based CNN on meshes, in particular, on semi-regular triangulated meshes. We showed that the convolution operation on semi-regular triangulated meshes has the property of translation, similar to that on the Euclidean space. The pooling operation on semi-regular triangulated meshes is analogous to that in the classic CNN in the Euclidean space. We employed this approach to the ADNI-2 data and compared its performance to that of the graph CNN \cite{Defferrard2016}. Our results showed that our vertex-based CNN algorithm was faster than the graph CNN. This is partly because the mesh coarsening procedure used for the pooling in the graph CNN \cite{Defferrard2016} requires adding fake vertices with the singletons such that each vertex has two children. This procedure increases the data dimension that is needed for  CNN. In contrast, the pooling operation in our proposed vertex-based CNN is with stride 2, similar to the downsampling factor achieved in the Euclidean space. Moreover, compared to the graph CNN, our proposed vertex-based CNN improved the classification accuracies of the five classifiers, except the EMCI and LMCI classifier. One of the potential limitations of our proposed approach is that it requires meshes to be semi-regular. In general, the construction of semi-regular meshes for medical image data is not an issue. However, our approach does not apply to graph data, such as social networks and citation networks and so on.

In the past decade, substantial studies reported the classification among CON, MCI, and AD based on the ADNI dataset (e.g., \cite{Liu_NIMG_2013,Nir_NBA_2015,Dyrba_NIMG_2015,Zhu_NIMG_2014,Yu_BSF_2016}. Some of them were based on multi-modal brain images and reported the classification accuracy better than that in Table \ref{tab2} (e.g., \cite{Nir_NBA_2015,Dyrba_NIMG_2015,Zhu_NIMG_2014,Yu_BSF_2016}). But the sample size was relatively small hence it is unclear on the robustness of the classification results. Nevertheless, our approach can be easily extended to multi-channel vertex-based CNN for handling multiple-modal or multiple structural data, such as diffusion properties of the cortex, cortical surface area and hippocampal shape. Compared to the existing studies based on cortical thickness ($~85\%$) \cite{Liu_NIMG_2013}, our approach reported the highest classification accuracy. To our best knowledge, our experiment employed the largest image data available in the ADNI-2 cohort, suggesting its potential robustness to other AD datasets.

\section{Ackowledgements}
We like to thank the National Supercomputing Centre Singapore for providing the computing resource for this study. The study was supported by Institute of Data Science at the National University of Singapore.

\bibliographystyle{splncs03}
\bibliography{references,intrinsic,deeplearning}

\end{document}